\documentclass[global,twocolumn]{svjour}
\usepackage{breakcites}

\usepackage{amsmath}
\usepackage{graphics}
\usepackage{multirow}
\usepackage{url}
\journalname{myjournal}

\usepackage{harvard}
\usepackage{inputenc}
\usepackage{xcolor}
\usepackage{subfloat}

\usepackage{tikz}
\usetikzlibrary{shapes,arrows,decorations.text}
\usetikzlibrary{trees}

\newcommand{\arcarrowdown}[8]
{\pgfmathsetmacro{\rin}{#1}
\pgfmathsetmacro{\rmid}{#2}
\pgfmathsetmacro{\rout}{#3}
\pgfmathsetmacro{\astart}{#4}
\pgfmathsetmacro{\aend}{#5}
\pgfmathsetmacro{\atip}{#6}
\fill[#7] (\astart:\rin) arc (\astart:\aend:\rin) -- (\aend+\atip:\rmid) -- (\aend:\rout) arc (\aend:\astart:\rout) -- (\astart+\atip:\rmid) -- cycle;
\path[decoration={text along path, text={|\Large\color{white}| #8}, text align={align=center}, raise=-0.5ex},decorate] (\astart+\atip:\rmid) arc (\astart+\atip:\aend+\atip:\rmid);
}

\newcommand{\arcarrowup}[8]
{ \pgfmathsetmacro{\rin}{#1}
\pgfmathsetmacro{\rmid}{#2}
\pgfmathsetmacro{\rout}{#3}
\pgfmathsetmacro{\astart}{#4}
\pgfmathsetmacro{\aend}{#5}
\pgfmathsetmacro{\atip}{#6}
\fill[#7] (\astart:\rin) arc (\astart:\aend:\rin) -- (\aend+\atip:\rmid) -- (\aend:\rout) arc (\aend:\astart:\rout) -- (\astart+\atip:\rmid) -- cycle;
\path[decoration={text along path,reverse path=true, text={|\Large\color{white}| #8}, text align={align=center}, raise=-0.5ex},decorate] (\astart+\atip:\rmid) arc (\astart+\atip:\aend+\atip:\rmid);
}

\begin{document}
\title{Using Simulation to Incorporate Dynamic Criteria into Multiple Criteria Decision Making}
\author{Uwe Aickelin \and Jenna Marie Reps \and Peer-Olaf Siebers \and Peng Li} 
\institute{School of Computer Science, University of Nottingham, Nottingham, NG8 1BB 
}
\date{Received: date / Revised version: date}
%
\maketitle
\begin{abstract}
In this paper we present a case study demonstrating how dynamic and uncertain criteria can be incorporated into a multi-criteria analysis with the help of discrete event simulation. The simulation guided multi-criteria analysis can include both monetary and non-monetary criteria that are static or dynamic, whereas standard multi-criteria analysis only deals with static criteria and cost benefit analysis only deals with static monetary criteria. The dynamic and uncertain criteria are incorporated by using simulation to explore how the decision options perform. The results of the simulation are then fed into the multi-criteria analysis. By enabling the incorporation of dynamic and uncertain criteria, the dynamic multiple criteria analysis was able to take a unique perspective of the problem. The highest ranked option returned by the dynamic multi-criteria analysis differed from the other decision aid techniques. The results suggest that dynamic multiple criteria analysis may be highly suitable for decisions that require long term evaluation, as this is often when uncertainty is introduced.

\end{abstract}
\section{Introduction}
Everyday we are faced with the task of making decisions. When a decision involves multiple criteria, the solution is often non-trivial \cite{doumpos2013} and it has been shown that humans struggle to make decisions when overloaded with information and choices \cite{iyengar2000}. As a consequence, methods have been developed that aid human decision making. In the context of decision making an option is a possible solution to the decision being made and the criteria are the set of considerations that are used to evaluate an option. In general, decision aid techniques score each option for each criterion and this is aggregated to determine the optimal option. 

If the options' scores for the criteria are monetary (i.e., quantitative), such as the cost of a TV, and are static, then methods such as cost effectiveness analysis (CEA) \cite{levin2001} or cost benefit analysis (CBA) \cite{svensson2010} can be used. Unfortunately, in many cases, options' scores for some of criteria are non-monetary (i.e., qualitative), for example you may `like' or `dislike' the style of a TV. It is difficult to directly compare monetary and non-monetary criteria scores, as they can have different scales. One method that has been developed to deal with decisions involving heterogeneous criteria scoring is the multi-criteria decision analysis (MCDA) method \cite{schroeder2011}. MCDA is a popular method for decision making, especially when data derived from human experts are involved \cite{gumus2016} as is also the case for our study of the port of Dover.

The MCDA method involves mapping the different criteria scores onto a comparable scale and then calculates a single overall score based on a weighting scheme provided by the stakeholders. Although MCDA can be used when there is a mixture of monetary and non-monetary criteria, it has a limited applicability when there is uncertainty \cite{lindhe2013}. In this paper we explore combining MCDA with simulation to overcome this shortcoming. We present a case study of implementing a multi-criteria decision making analysis that using simulation to determine some of the performances measures that are uncertain.  We consider the expansion of the port of Dover and use simulation to estimate the flow of traffic through to port for three scenarios (specified by the port authorities). 

In this paper we present a case study to illustrate the decision differences when only considering; a) static monetary criteria, b) static monetary/non-monetary criteria and c) static/dynamic monetary/non-monetary criteria. The case study investigates the optimal expansion strategy for the port of Dover, taking into consideration the cost of the expansion (including additional staff and greenery to reduce the impact of increased CO$_{2}$ emissions) and the customers' satisfactions (dependent on their initial mood, whether they travel alone and how long they spend queuing). The objective of this paper is to investigate the difference in the final optimal decision between CBA, static MCDA (only considering static criteria) and dynamic MCDA (when dynamic criteria are considered). The dynamic MCDA incorporates simulation to give insight into the dynamic criteria. The differences can then be used to try and identify decision situations that may benefit from incorporating dynamic criteria.

The results give insight into understanding which technique is most suitable for aiding a given decision problem (i.e., when should you incorporate non-monetary criteria or simulation?). The novelty of this paper is that we present a case study that explores applying a discrete event simulation (DES) to generate values for the non-static variables that are then fed into a MCDA along with static variables. 

The continuation of this paper is as follows. In section \ref{bk} we give an overview of cost benefit analysis and multi-criteria decision analysis. This is followed by a case study in section  \ref{probf}. We then implement the decision analysis and simulation for the decision of expanding the port of Dover in section\ref{sim_mcda}. Section \ref{des} presents the results of the case study and a discussion and the paper concludes with section \ref{conc}.

\section{Background} \label{bk}

\subsection{multi-criteria Decision Analysis}
Cost Benefit Analysis is decision aiding tool used when only static monetary values are considered. The overall benefit of each option is compared with the overall cost of each option to calculate the net score. For the i\textsuperscript{th} option the net score is calculated as,
\begin{equation}
N_{i}= \sum_{j} C_{ij} - \sum_{k} B_{ik}
\end{equation}
Where $C_{ij}$ is the j\textsuperscript{th} cost of the i\textsuperscript{th} option and $B_{ik}$ is the k\textsuperscript{th} benefit of the i\textsuperscript{th} option. In general, the option with the greatest net gain is chosen. CBA has been used to aid decision making in numerous fields including transportation \cite{eliasson2009} or transportation infrastructure investments \cite{damart2009}. The main advantages of the CBA are its simplicity, meaning it is easy to understand, and the fact that is takes the net score into consideration. Therefore, the option of doing nothing will be chosen if every other option has a negative net score. In other methods such as cost effectiveness analysis, this is not always the case. 

MCDA is a special case of CBA - The goal of the MCDA is to aid decision making when the criteria have difference scales (monetary and non-monetary) and have various importance on the final decision. It is important to identify the objective and measure the score of each option for addressing the objective on a chosen set of criteria. These all depend on the decision makes and stakeholders, as the criteria importance and scores can be subjective. The MCDA combines all of this information to generate a single value for each option that can be used to compare the options and chose the more suitable one. Due to the subjectivity of measures used in the analysis, sensitivity analysis is often applied to see how stable the MCDA is. The MCDA generally follows the steps presented in Figure \ref{mcda}, this is the order that is followed in the case study section. 

A popular and frequently implemented method to solve a MCDA problem is the Analytical Hierarchical Process (AHP) \cite{saaty1990}. The AHP method determines weights to assign to each criteria, based on stakeholders preferences, and uses these to calculate an overall single score for each decision option. The options are then ranked by the overall score, and the option obtaining the greatest score is recommended. 

Formally, denoting the i\textsuperscript{th} option by $A_{i}$ and the j\textsuperscript{th} criterion by $C_{j}$, then the performance of option $i$ for criterion $j$ is $a_{ij}$. This can be represented as a matrix,
\begin{equation}
A= \bordermatrix{~ & \mbox{C$_{1}$} & \mbox{C$_{2}$} & ... & \mbox{C$_{n}$} \cr
\mbox{A$_{1}$} & a_{11} & a_{12} & ... & a_{1n} \cr
\mbox{A$_{2}$} & a_{21} & a_{22} & ... & a_{2n} \cr
. & . & . & ... & . \cr
\mbox{A$_{m}$} & a_{m1} & a_{m2} & ... & a_{mn} \cr}
\end{equation}
The weights assigned to each criterion (based on stakeholder preferences) can be presented by the vector,
\begin{equation}
W= [w_{1}, w_{2}, …, w_{n}]
\end{equation}
where $w_{j}$ is the weight assigned to the j\textsuperscript{th} criterion, this is the importance of the criterion to the stakeholders. The vector containing the AHP measure for each option is calculated by,
\begin{equation}
AW^{T}= [\sum_{k} a_{1k}w_{k},\sum_{k} a_{2k}w_{k},..., \sum_{k} a_{mk}w_{k} ]^{T}
\end{equation}
and the optimal option is chosen by,
\begin{equation}
\arg \max_{i} \sum_{k} a_{ik}w_{k}
\end{equation}

MCDA has been applied for transportation and in \cite{tudela2006} the option ranking of MCDA was compared with CBA when considering a case study involving transport investment. This case study showed that the decision option ranking differed between MCDA and CBA when non-monetary criteria were incorporated. In the AHP calculation, the options' criteria scores are assumed to be known, but if this is not the case, difficulties arise in calculating the measure of overall weighted criteria scores.

\subsection{Simulation and related work}

A simulation serves to replicate the behaviour of a real world process or system over time. This often involves constructing a model that is an abstraction of the real world system and running it over time. The simulation can then be used as a means to understand the underlying dynamics of the process. There are various techniques for simulation, some are continuous \cite{ford1999}, other are discrete \cite{cassandras2008}. Discrete event simulation (DES) considers a sequence of events over time where the state changes occur discretely. DES can be stochastic or deterministic and is often used when modelling human flow \cite{hung2007} \cite{beck2011} \cite{gill2016}. Simulation is becoming a more and more popular decision aid tool, with increasingly closer links to business performance measures such as Key Performance Indicators \cite{jaha2017} and many successful applications to port scenarios \cite{li2016}.

The most closely related work to ours is the paper by Meng et al \cite{meng2014} where the authors use discrete event simulation to aid an Analytic Hierarchy Process multi-criteria analysis. They consider a case study for the problem of vegetable grafting operation design. The performance of a grafting operation is affected by complex factors such as weather and disease outbreaks so the authors use simulation to approximate the performance.  As mentioned earlier, one of MCDA's disadvantages is its difficulty to handle uncertainty. When a criterion's score is uncertain/dynamic it can sometimes be inferred using methods such as hedonic price techniques \cite{boardman2010} (the relationship of the criterion on a product's monetary value is used to infer its value) or by implementing surveys. However, these techniques are not always possible or may be too expensive.

To address this issue, some researchers have used simulation as a tool to explore uncertain/dynamic criteria scores \cite{meng2014}.A simulation can be run for a particular scenario to estimate each option's score when considering the dynamic criteria, these scores can then be fed into an MCDA to enable the inclusion of dynamic criteria. By combining simulation and MCDA, decisions can be made based on a mixture of static and dynamic criteria that can have monetary or non-monetary scores. For example, consider the situation that a company wants to open a new store and has the choice of three different locations. The company may choose to consider the cost of building the store, the yearly profit and local impact to the community. As none of the stores is built, it is not possible to know each store's exact yearly profit or impact to the community. With unknowns, the MCDA cannot be applied and this is where simulation comes to the rescue. Simulation alone has often been used as a tool to aid decision making in a variety of fields including healthcare \cite{karnon2003} and the environment \cite{muys2010}. It is an useful tool when the criteria scores are dynamic (i.e., uncertain) for example criteria involving future predictions or where precise measurements are unavailable due to constraints including time or money.

Another similar paper is by Brito et al \cite{brito2010} that described a case study of using simulation to aid the decision making for steel plant logistic system planning.  The paper demonstrated the usefulness of combining discrete event simulation and multi-criteria analysis for ranking different scenarios when there is uncertainty. \cite{bale2017} uses a combination of MCDA and Monte Carlo simulation to compare different energy generation scenarios and choose the most appropriate. Also in \cite{sherman2013}, the authors compared a selection of decision aiding techniques that can enable cost-benefit analysis to be implemented when the cost is uncertain. They implemented a case study investigating cargo screening at Calais port and applied scenario analysis, decision trees or simulation to estimate the cost. The results showed that the CBA aided by simulation returned unique results. In \cite{chan1996} the authors use simulation to include uncertain criteria into the Analytic Hierarchy Process for the manufacturing systems case study.These examples provide evidence to suggest using simulation to aid a decision making algorithm that considers criteria that are dynamic and uncertain can give new insight. However, the simulation adds complexity to the decision making process and often requires additional assumptions, therefore it may not always be useful.

\subsection{Our combined simulation and MCDA approach}
The above papers all demonstrated the usefulness of using simulation to approximate performance measures that are uncertain due to the complexity of the system, e.g. by enabling the incorporation of dynamic criteria \cite{meng2014}.This motivated us to use the same methodology for the port of Dover study, as the traffic flow and customer satisfaction are affected by many unexpected events. There has been no other case study considering decision making for vehicle flow through ports and this work was motivated by the real life problem of how the authorities of Dover could expand to prepare for increased future traffic. 

The literature shows that some researchers have combined simulation and AHP by running the simulation to investigate the estimated score of each option for a specific dynamic criterion. The results of the simulation are then fed into the MCDA \cite{chan1996}. Alternatively, others have applied a two step approach that firstly implements AHP to reduce the number of options by selecting those ranked highest and then applies simulation to choose between these remaining options \cite{li2009}. In this paper the first approach is implemented as it is more appropriate to the structure of our case study.

\begin{figure} \centering
\caption{A flowchart of the MCDA process.}\label{fig:fig0}
\label{mcda}
\begin{tikzpicture}[auto]
\tikzstyle{block} = [rectangle, draw=white, fill=black!30,text width=10em, text centered, rounded corners, minimum height=3em, text=white]
\tikzstyle{block_add} = [rectangle,dashed, draw=black, fill=white!30,text width=8em, text centered, rounded corners, minimum height=3em, text=black]


\fill[even odd rule,black!30] circle (3.8) circle (2.2);
\draw[color=black] (0,0) circle (3) node {\Large\bf Stakeholders};
\arcarrowup{2}{3}{4}{60}{115}{-10}{black,draw=black!50!black,very thick}{Options}
\arcarrowup{2}{3}{4}{0}{55}{-10}{black,draw=black!50!black,very thick}{Criteria}
\arcarrowdown{2}{3}{4}{300}{355}{-10}{black,draw=black!50!black,very thick}{Simulation}
\arcarrowdown{2}{3}{4}{240}{295}{-10}{black,draw=black!50!black,very thick}{Weights}
\arcarrowdown{2}{3}{4}{180}{235}{-10}{black,draw=black!50!black,very thick}{Score}
\arcarrowup{2}{3}{4}{120}{175}{-10}{black,draw=black!50!black,very thick}{Examine}

\end{tikzpicture}
\end{figure}
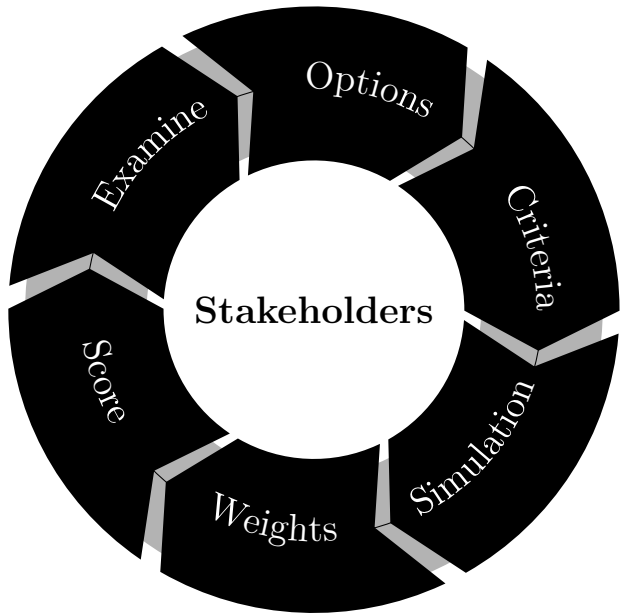

Hence the first step of our combined MCDA and simulation approach is identifying the context of the decision; this involves identifying the decision makers and key players (these are collectively known as the stakeholders) in addition to identifying the aims of the decision. This is then used to develop a selection of options (possible solutions based on simulation runs). Finally we need to select the criteria used to evaluate each option, calculate an overall score and report back. A summary of our approach is displayed in Figure \ref{fig:fig0}. 

\section{Case Study: Port of Dover} \label{probf}
\subsection{The decision context} 
In this case study we focus on the question of what is the optimal infrastructure expansion strategy when considering service, safety and income? The decision makers in this problem are Dover Harbour Board and the key players are the staff working at the port, the Department of Transport, the local community surrounding the port and the general public. The collection of decision makers and key players are referred to as the stakeholders. 

Dover is a major ferry port connecting the UK and France. In 2016, $4 808 010$ vehicles passed through the port \cite{dover} and the Dover Harbour Board of directors are expecting the vehicle flow to increase annually. At present, the Dover Harbour Board of directors is aiming to double the traffic over the next ten years. This corresponds to a desired annual growth averaging 7\% but the actual annual growth is uncertain. After discussions with the Board of directors the best predictions for the annual growth is a range between 0\% and 20\% each year. The Board of directors also expect that the ratio of lorry to non-lorry vehicles will increase (i.e., the increase in lorry traffic will be greater than the increase in non-lorry traffic). If the travel flow increases as predicted then at some point in the future the current road network within the port may not be able to cope with the increased traffic and this may lead to long queues and unhappy customers. If the customers of the port become unhappy with the service, they may use alternative means to travel in the future and this would negatively impact the port's future success. Therefore the aim of this case study is to identify the optimal expansion strategy to implement such that the port of Dover can cope with the additional traffic, ensuring customer satisfaction while maximising income.

Figure \ref{fig:fig1} presents the plan of the port. The route taken and main landmarks are highlighted. The first milestone of the customers' journeys is the passport checking. After the passport checking the vehicles then move towards the weighbridge, there are two lanes joining the passport checkpoint and the weighbridge. The weighbridge point then splits into seven lanes, five lanes are for the lorry users to be weighed and the remaining two are for non-lorry customers (as they are not weighed). The weighbridge is a common place for queuing to occur and queues forming at this point will also lead to queues further down the route, such as before the passport check. 

\begin{subfigures}
\begin{figure*}\centering
\caption{The satellite view of the Port of Dover with the key points marked.}\label{fig:fig1}
\includegraphics[width=0.8\textwidth]{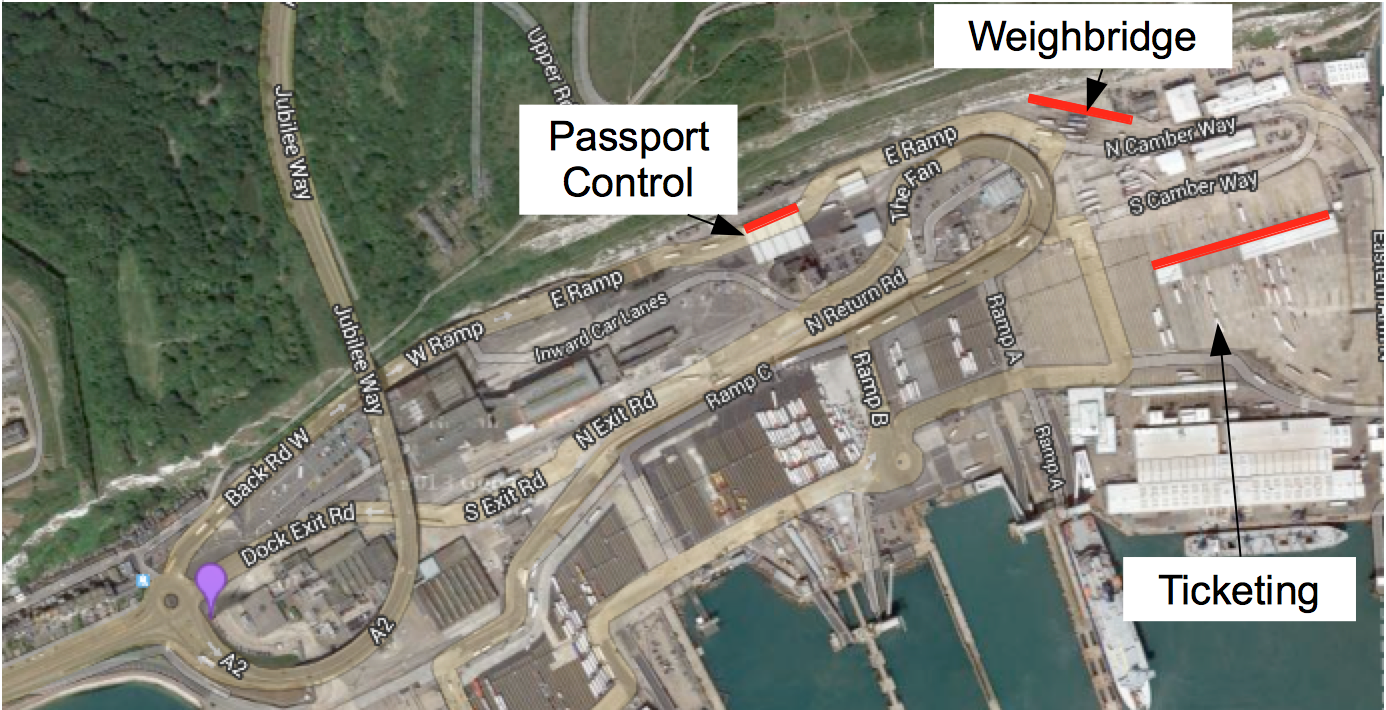}
\end{figure*}

\begin{figure*}\centering
\caption{The plan of the Port of Dover with the passport control in green and the weighbridge and ticketing in blue.}
\label{fig:fig2}
\includegraphics[width=0.8\textwidth]{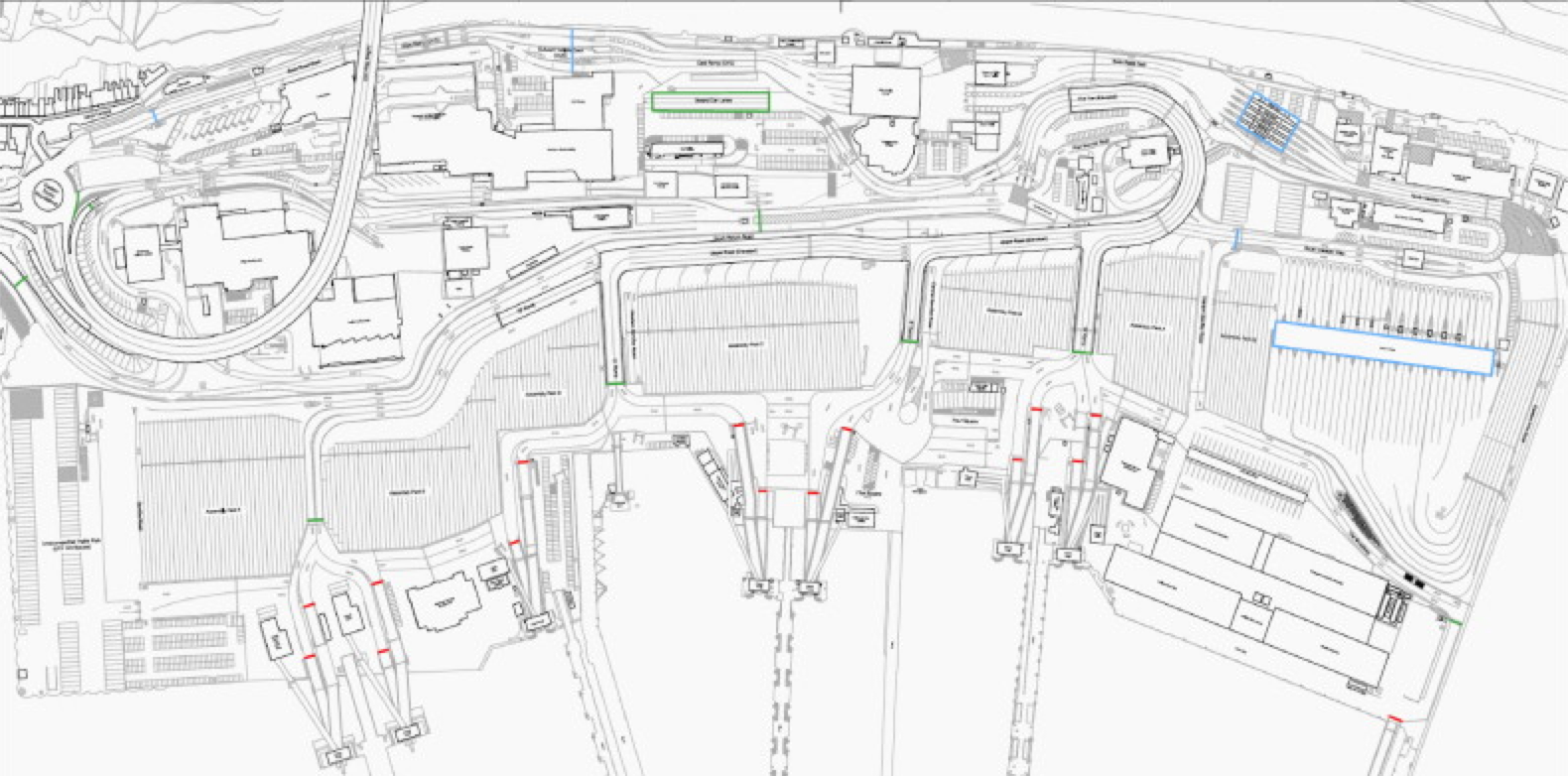}
\end{figure*}

\end{subfigures}

\subsection{Scenarios}
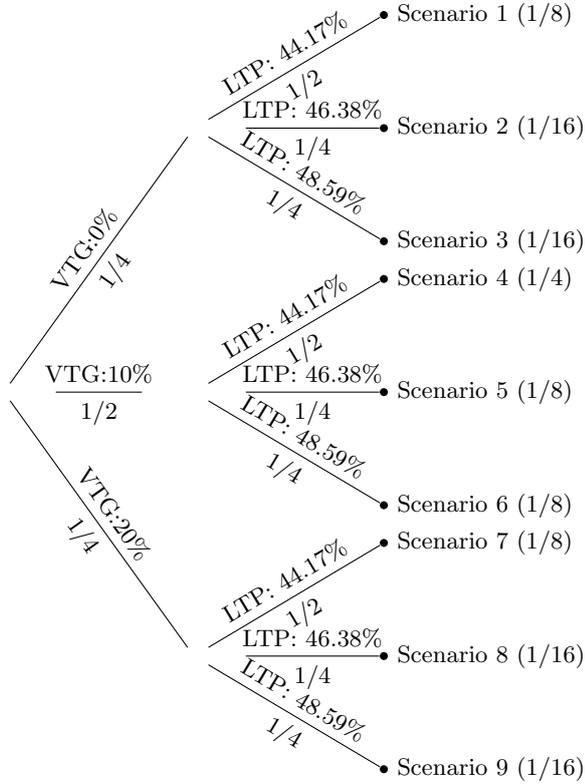
\begin{figure}\centering
\caption{The probability tree of the nine different scenarios of the case study when considering the three possible vehicle traffic grows (VTGs) and the three possible lorry traffic percentages (LTPs).}
\label{tree}
\tikzstyle{level 1}=[level distance=2.5cm, sibling distance=3.5cm]
\tikzstyle{level 2}=[level distance=2.5cm, sibling distance=1.5cm]

\tikzstyle{bag} = [text width=3.5em, text centered]
\tikzstyle{end} = [circle, minimum width=3pt,fill, inner sep=0pt]

\begin{tikzpicture}[grow=right, sloped]
\node[bag] {}
child {
node[bag] {} 
child {
node[end, label=right:
{Scenario 9 (1/16)}] {}
edge from parent
node[above] {LTP: 48.59\%}
node[below] {$1/4$}
}
child {
node[end, label=right:
{Scenario 8 (1/16)}] {}
edge from parent
node[above] {LTP: 46.38\%}
node[below] {$1/4$}
}
child {
node[end, label=right:
{Scenario 7 (1/8)}] {}
edge from parent
node[above] {LTP: 44.17\%}
node[below] {$1/2$}
}
edge from parent 
node[above] {VTG:20\%}
node[below] {$1/4$}
}
child {
node[bag] {} 
child {
node[end, label=right:
{Scenario 6 (1/8)}] {}
edge from parent
node[above] {LTP: 48.59\%}
node[below] {$1/4$}
}
child {
node[end, label=right:
{Scenario 5 (1/8)}] {}
edge from parent
node[above] {LTP: 46.38\%}
node[below] {$1/4$}
}
child {
node[end, label=right:
{Scenario 4 (1/4)}] {}
edge from parent
node[above] {LTP: 44.17\%}
node[below] {$1/2$}
}
edge from parent 
node[above] {VTG:10\%}
node[below] {$1/2$}
}
child {
node[bag] {} 
child {
node[end, label=right:
{Scenario 3 (1/16)}] {}
edge from parent
node[above] {LTP: 48.59\%}
node[below] {$1/4$}
}
child {
node[end, label=right:
{Scenario 2 (1/16)}] {}
edge from parent
node[above] {LTP: 46.38\%}
node[below] {$1/4$}
}
child {
node[end, label=right:
{Scenario 1 (1/8)}] {}
edge from parent
node[above] {LTP: 44.17\%}
node[below] {$1/2$}
}
edge from parent 
node[above] {VTG:0\%}
node[below] {$1/4$}
};
\end{tikzpicture}
\end{figure}

The values for the future vehicle traffic and percentage of the total vehicles going through the port that are lorries are uncertain. To deal with this uncertainty we consider the percentage annual vehicle traffic growth (VTG) and percentage lorry traffic percentage (LTP) to be discrete random variables and implement a discrete probabilistic model. Due to the relationship between the VTG and LTP being unknown, we assume they are independent. The discrete probabilistic model considers the probability of various potential future scenarios. These scenarios were developed based on discussions with the Dover Harbour Board and by analysing historical data. Three possible values for the VTG over the next year are chosen, $S(VTG)=\{ 0, 10, 20 \}$ with respective probabilities of $P(VTG= 0)=0.25$, $P(VTG= 10)=0.5$ and $P(VTG= 20)=0.25$. Based on the figures extracted from the annual reports, the percentage of vehicles entering the port that are lorries over the 2006-2011 time period have ranged between 41.86\%-45.78\% with an average percentage of 44.17\%. During times of economic weakness the number of lorries travelling though the port decreases, however, as the economy is recovering the Dover Harbour Board expect the percentage to range between 44.17\% and 48.59\% (a 10\% increase) over the next year. This prompts the derivation of support of the LTP variable, $S(LTP)=\{ 44.17, 46.38 ,48.59 \}$ with respective probabilities of $P(LTP= 44.17)=0.5$, $P(LTP= 46.38)=0.25$ and $P(LTP= 48.59)=0.25$. There are therefore nine different future scenarios (different combination of possible VTG and LTP values), these are presented in Figure \ref{tree}. 

\subsection{The options}
After discussions with the Dover Harbour Board of directors three options for the expansion were identified;
\begin{enumerate}
\item The first option was to do nothing.
\item The second options is to expand the road network by relocating offices next to the weighbridge lanes to make space for an additional lorry lane at the weighbridge. 
\item The third option is to expand the road network by relocating offices next to the weighbridge lanes to make space for an additional non-lorry lane at the weighbridge.
\end{enumerate}
There is only room for one additional lane in the space currently occupied with office building, so this limited the number of options available.

\section{Combined Simulation and MCDA Approach} \label{sim_mcda}
\subsection{The criteria }\label{criteria}
Our combined approach as described in Figure \ref{fig:fig0} first requires us to establish the MCDA criteria.The values presented below have been based on the 2010-2011 financial report \cite{dover}. The decision criteria are:
\begin{description}
\item[C] Costs
\begin{description}
\item[C.1] Finance - Any costs spent on expanding the port will come out of savings and therefore there will be a loss from interest.
\item[C.2] Environment - The increase in traffic/queuing may lead to additional carbon emissions and this will need to be addressed. In the case of additional carbon dioxide levels new greenery will be planted. 
\item[C.3] Safety - The current accident incident rate at the port of Dover is 1.2 (compared with 1.9 on average for other ports). To maintain this rate new staff and equipment (i.e., a security car) will be required when there is an increase in traffic. It was decided that for every 10\% increase in traffic five additional members of staff will be hired and one new car will be purchased. 
\item[C.4] Facility Building - The cost of relocating the offices to make space for a new lane. 
\end{description}
\item[B] Benefits
\begin{description}
\item[B.1] Benefits for the Port - The short term benefits considered are traffic increase and profit increase. The long-term benefits that would result from a successful expansion are potential consultation work, new port partnerships and environment protection at the port. 
\item[B.2] Benefits for the Customers - This includes the patients' safety, the patients' satisfaction with nearby customers and the frequency of queuing within the port (specifically at the weighbridge).
\item[B.3] Benefits for the Local Community - This includes the creation of jobs due to the expansion and the increase in trade around the port as a consequence of vehicle traffic growth.
\end{description}
\end{description}

\subsection{Scoring options for criteria} \label{scoring}
For the cost scoring we take into consideration the loss of interest due to spending savings on the expansion. Based on the 2010-2011 financial report, the Board earned \textsterling $1.031$ million on a cash asset of \textsterling$46.092$ million, this corresponds to an interest rate of approximately 2.39\%, therefore,
\begin{equation}
INT = 1+ 1031000/46 092 000
\end{equation}
The cost of a new lane including the office building relocation is \textsterling$90000$, this value is independent of the VTG or LTP. The effective cost, including the loss of interest, is therefore \textsterling$92148.8$. 

The cost of additional security staff to maintain the current level of driving safety when there is additional traffic is dependent of the VTG. The Board feel that 5 new members of staff and one car is required for every 10\% increase in VTG. The total cost of salaries over 2010-2011 was \textsterling$12.488$ million for 344 members of staff. This includes a \textsterling$0.502$ million remuneration for 8 key members of staff, so the average will be adjusted. The new car will cost \textsterling$20000$ and have an annual maintenance fee of \textsterling$2000$. The cost for every 10\% increase in VTG is,
\begin{equation}
\begin{split}
C_{s} &= (5 \times \mbox{average cost staff} + \mbox{cost of car} )\times \mbox{interest}\\
&= (5 \times\frac{1248800-502000}{344-8} + 22000)\times INT\\
&= (178363 + 22000)\times INT\\
&= 205146.8\\
\end{split}
\end{equation}
Therefore, the expected cost for each option is,
\begin{equation}
\begin{split}
E(C_{s}) =& P(VTG=0) \times 0 + P(VTG=10) \times 200841.87 \\
&+ P(VTG=20)\times 401683.74 \\
=& 0.25\times 0 + 0.5\times 205146.8 + 0.25 \times 410293.6 \\
=& 205146.8 \\
\end{split}
\end{equation}
The cost of planting trees to maintain the environmental conditions by combating carbon emissions is also dependent on the VTG. The Board feel that any increase in carbon emissions will require planting greenery at a cost of \textsterling $50000$. The carbon emissions will depend on the traffic and the speed of the traffic. We represent the current consumption of gasoline for each car passing through the port by 1 consumption unit and the current vehicle traffic by 1 traffic unit. The total consumption at a given vehicle traffic is the consumption unit multiplied by the traffic unit, so this is currently 1 ($1 \times 1$). However, due to the additional lane increasing the traffic flow, for options 2 and 3, we consider each car represents 0.85 consumption units. If the total consumption is greater than $1$, then greenery is required costing \textsterling$50 000$ (effectively \textsterling$51193.78$), therefore, the cost of maintaining the environment is,
\begin{equation}
C_{e} = \left \{ \begin{array}{ll} 51193.78 \times H({(1 (1+VTG))-1)} & \quad \mbox{ option 1} \\
$51193.78$ \times H({(0.85(1+VTG))-1)} & \quad \mbox{otherwise} \\
\end{array} \right.
\end{equation}
where $H(x)$ is the heaviside function that is $1$ if $x>0$ and $0$ otherwise. The expected environmental cost for option 1 is,
\begin{equation}
\begin{split}
E(C_{s}) =& P(VTG=0) \times 0 + P(VTG=10) \times 51193.8 \\
&+ P(VTG=20)\times 51193.8 \\
=& 0.25\times 0 + 0.5\times 51193.78 + 0.25 \times 51193.8 \\
=& 38395.3 \\
\end{split}
\end{equation}
and the expected environmental costs for option 2 and 3 are,
\begin{equation}
\begin{split} 
E(C_{s}) =& P(VTG=0) \times 0 + P(VTG=10) \times 0 \\
&+ P(VTG=20)\times 51193.8 \\
=& 0.25\times 0 + 0.5\times 0 + 0.25 \times 51193.8 \\
=& 12798.5 \\
\end{split}
\end{equation}

The benefits of each option in terms of profit are the dependent on the traffic passing through the port. In 2010-2011 the profit was \textsterling$7.588$ million and we assume that the profit growth has a linear relationship with the traffic growth. Therefore, the additional profit is,
\begin{equation}
B_{p} = VTG \times 7588000
\end{equation}
and the expected profit for each option is,
\begin{equation}
\begin{split} 
E(B_{p}) =& P(VTG=0) \times 0 + P(VTG=10) \times 758800 \\
&+ P(VTG=20)\times 1517600 \\
=& 0.25\times 0 + 0.5 \times 758800 + 0.25 \times 1517600 \\
=& 758800 \\
\end{split}
\end{equation}

\begin{figure*}\centering
\caption{The concept model of the simulation.}\label{fig:fig01}
\includegraphics[trim=0cm 11.2cm 0cm 1cm, clip=true,width=0.95\textwidth]{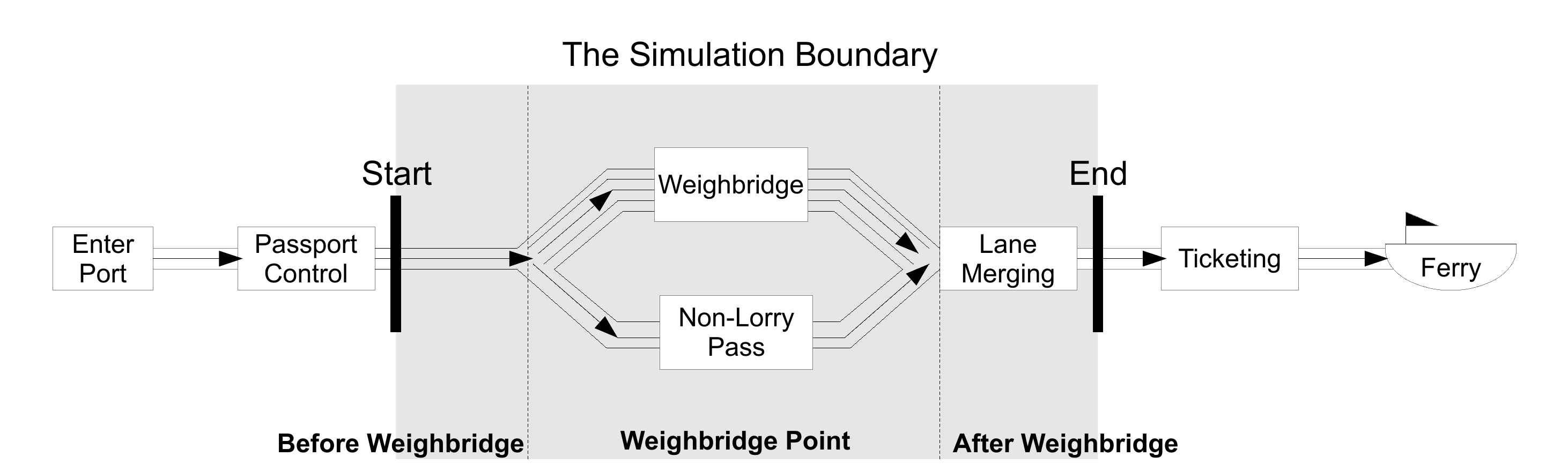}
\end{figure*}
\begin{figure*}
\centering
\caption{Graphical representation of the conceptual model.}
\label{sat}
\includegraphics[trim=0cm 11.5cm 0cm 0cm, clip=true,width=\textwidth]{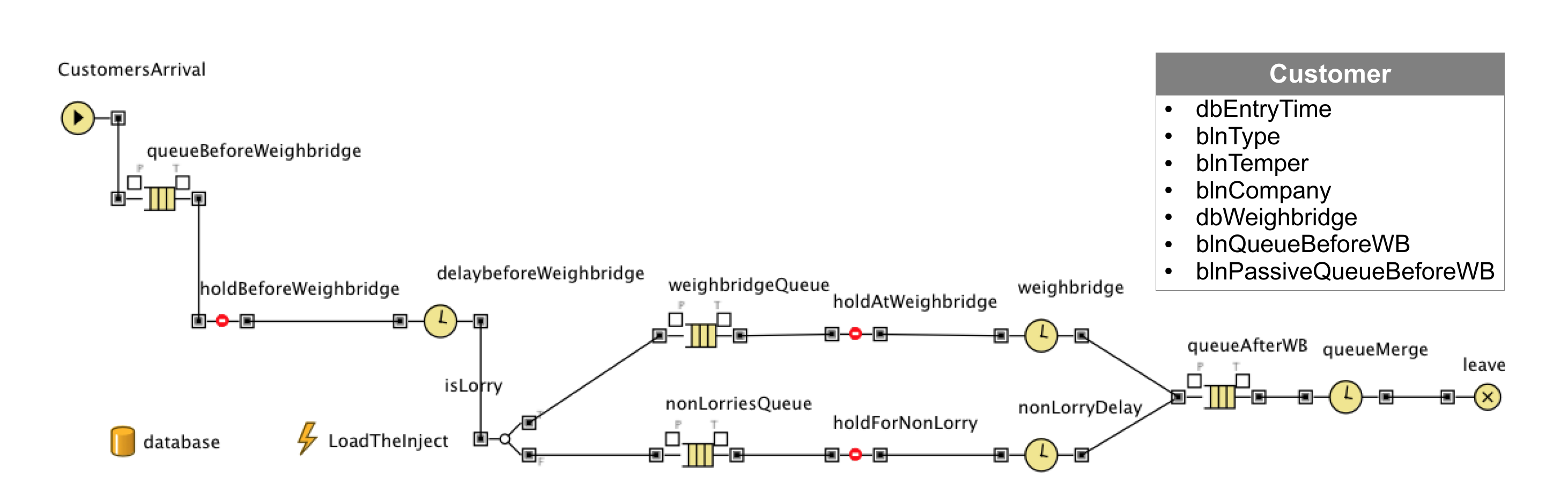}
\end{figure*}
\begin{table*} \centering
\caption{How the customer entity dissatisfaction were derived based on the simulation.  Each table cell corresponds to the percentage of how dissatisfied a customer is based on their simulated experience.}
\label{tab:cs}
\begin{tabular}{cccc}
& No Queuing & Non-passive Queuing & Passive Queuing \\ \hline
Good temperament \& Not alone & 0 & 20 & 50 \\
Good temperament \& Alone & 0 & 50 & 80 \\
Bad temperament \& Not Alone & 0 & 40 & 70 \\
Bad temperament \& Alone & 0 & 70 & 100 \\
\end{tabular}
\end{table*}

The remaining criteria are non-monetary. The non-monetary criteria that are known are the road safety, local profits and job opportunities. All three options have the road safety priced into them to ensure the safety of customers, so they all cover this criterion. The local profit is dependent on the VTG, as a higher VTG means more customers passing through the local area, the expansion strategy is unlike to impact the short term VTG, so all three options will have a similar VTG and therefore a similar benefit to local profits. Options 2 and 3 require additional staff to build the extension and to ensure the roads remain safe, however, option 1 does not require hiring any additional staff, so only options 2 and 3 benefit job opportunities.

\subsection{Simulation} 
The customer satisfaction at the different $VTG$ possibilities is dynamic and unknown so we implement an object oriented DES to score the related criteria. The simulation model was constructed based on interviews with Dover's transportation system experts. The concept model of the simulation is displayed in Figure \ref{fig:fig01}. 

The model parameters are:
\begin{itemize}
\item Customers' arriving rate (dependent on VTG)
\item Number of weighbridges (dependent on option)
\item Number of non-lorry lanes at weighbridge
\item Size of the queue at the weighbridge lorry lanes
\item Size of the queue at the weighbridge non-lorry lanes
\item Service time at the weighbridge 
\item Number of lorry vehicles (LTP)
\item Percentage of customers driving alone
\item Percentage of customers with a good temperament.
\end{itemize}

The components of the model are:
\begin{itemize}
\item Customers 
\begin{itemize}
\item Arrival time - this is generated based on historical data
\item Customer type (lorry or non-lorry)- this is generated based on historical data
\item Temper - a customer has a 10\% chance of having a bad temper before enter the port
\item Company - a customer has a 10\% chance of travelling with others
\item Queuing before weight bridge point
\item Passive queuing before weight bridge point
\end{itemize}
\item Queues
\begin{itemize}
\item Capacity
\end{itemize}
\item Roads
\begin{itemize}
\item Travel time (distribution approximated using historical data)
\end{itemize}
\end{itemize}

The focus of the simulation is the weighbridge, therefore rather than simulating the whole flow through the port, the simulation only considers three parts; before the weighbridge, at the weighbridge point and after the weighbridge. The model starts at the point before the weighbridge. If there is available space at the weighbridge point (a free capacity of 2 or more at the weighbridge point queues) then the customer will advance otherwise they will join an unlimited capacity queue before the weighbridge. The queue is assumed to be unlimited due to the length of road between the port entrance and this point. The travel time before reaching the weighbridge point is represented by a delay. 

The customer then enters the weighbridge point and if their vehicle is a lorry then they go through the lorry lanes otherwise they go through the non-lorry lanes. If there is available space at the merging point then the customer precedes, otherwise they join their respective queue at the weighbridge point (queue for weighbridge if lorry or queue for non-lorry pass if non-lorry). The queues at the weighbridge point have a capacity that corresponds to the number of lanes. The service and travel time through the weighbridge point for lorries are combined and modelled using a single delay and the travel time through the weighbridge point for non-lorries is also modelled using a delay. If the customer is type lorry then the service and travel time though the weighbridge is normally distributed \cite{Roadknight2012} with a mean of 80 seconds and standard deviation of 2 seconds. If the customer is type non-lorry then the time it takes to go through the weighbridge point is normally distributed with mean 27 seconds and standard deviation 2 seconds. After merging, the customer leaves the simulation. The merging point queue has a capacity of 2 and the time it takes to merge is represented by a delay. In our model the customer enters a queue before entering the weighbridge point when the weighbridge lane and non-lorry lane have a free capacity of less than 2. The customers queue at the weighbridge point when there is no free capacity at the merging point.

In this simulation we assume that only queuing prior to the weighbridge impedes the satisfaction of a customer. A customer meets passive queuing when he enters a queue formed at the point before the weighbridge that was caused by a vehicle type opposite to the type he is driving. Passive queuing leads to the customer being more dissatisfied. For each customer entity within the simulation, his dissatisfaction with the journal is determined based on whether he is travelling alone, has a bad temperament or meets queuing. A customer's temperament is randomly assigned and a bad temperament occurs with a probability of 10\%. Whether a customer travels alone or with company is also randomly assigned with customers having a 90\% probability of travelling alone. These probabilities were derived using historical data. Table \ref{tab:cs} displays how each customer's dissatisfaction is calculated based on the simulation results.

Experts experimenting with the model and changing parameter values verified the model. Each constituent was also checked to ensure it was implemented correctly. In addition, historical data collected via CCTV at the entrance point of the weighbridge over 48 hours between 6/2/2010 - 8/2/2010 \cite{Roadknight2012} was used to validate and verify the simulation accurately represents the vehicle flow within the port. The model, presented in Figure \ref{sat}, was implemented in Anylogic 6.8.1 University Edition. The simulation was run for 365 days but to prevent initialisation bias the first 20 days were ignored. All simulations were repeated 10,000 times and mean values are reported.

The arrival rate of vehicles was assumed to be uniformly distributed over time and was calculated using historical traffic volume recorded over the year. Based on the 48080010 vehicles entering and leaving the port over 2010-2011 we assume half of them entered the port (2404005) and this corresponds to 4.57 vehicles arriving every minute. 

Historical data shows that the peak time for the lorry traffic is between 12:00 and 18:00 and during this time the lorry traffic percentage is generally 75\%. Therefore, in the simulation the temporal lorry traffic percentage ($tLTP$) is,
\begin{equation}
tLTP(t)= \left\{ 
\begin{array}{l l} 75 & t \in [12,18] \\
(LTP-75 \times 6/24)(24/18) & \mbox{ otherwise}
\end{array} \right.
\end{equation}
This ensures that the daily average lorry traffic percentage is equal to the $LTP$, but incorporates the knowledge of a higher rate between 12:00 and 18:00. 

The results of the simulation when run for the different scenarios and options are presented in Table \ref{simu}. The expectations for the queuing frequency, passive queuing frequency and bad temperament frequency for each strategy was found by summing the product of the probability of the scenario and frequency over the nine scenarios.  Historical data shows that the peak time for the lorry traffic is between 12:00 and 18:00 so the percentage of traffic between 12:00 and 18:00 is calculated as $LTR \times 4 \times 0.426$ and the percentage of traffic outside of these hours is calculated as $LTR \times 4/3 \times (1-0.426)$. For example, when the daily $LTR$ is 44.15\% then the percentage of lorry between 12:00 and 18:00 is 75\% and the percentage outside this time is 33.9\%, averaging the 44.15\%. The results of the simulation when run for the different scenarios and options are presented in Table \ref{summ_score}. The expectations for the queuing frequency, passive queuing frequency and bad temperament frequency for each strategy was found by summing the product of the probability of the scenario and frequency over the nine scenarios. 

\begin{table}[t!]
\caption{The probability in \% of a customer queuing (Queue), meeting passive queuing (pQueue) or having a bad temperament (Angry) for the different scenarios and options.}
\label{simu}
\begin{tabular}{cccc} \hline
\multicolumn{4}{c}{Option 1} \\ \hline
Scenario & Queue & pQueue & Angry \\
1 (VTG:0, LTP:44.17) & 0 & 0 & 0 \\
2 (VTG:0, LTP:46.38) & 0.01 & 0.003 & 0.008 \\
3 (VTG:0, LTP:49.59) & 0.72 & 0.14 & 0.40 \\
4 (VTG:0.1, LTP:44.17)& 1.31 & 0.35 & 0.75 \\
5 (VTG:0.1, LTP:46.38)& 6.86 & 1.75 & 3.88 \\
6 (VTG:0.1, LTP:49.59)& 14.92 & 3.69 & 8.42 \\
7 (VTG:0.2, LTP:44.17)& 17.82 & 5.48 & 10.37 \\
8 (VTG:0.2, LTP:46.38)& 26.86 & 8.21 & 15.6 \\
9 (VTG:0.2, LTP:49.59)& 34.55 & 10.62 & 20.13\\ \hline
Expected overall & 9.16 & 2.64 & 5.28 \\ \hline
\multicolumn{4}{c}{Option 2} \\ \hline
Scenario & Queue & pQueue & Angry \\
1 (VTG:0, LTP:44.17) & 0 & 0 & 0 \\
2 (VTG:0, LTP:46.38) & 0 & 0 & 0 \\
3 (VTG:0, LTP:49.59) & 0 & 0 & 0 \\
4 (VTG:0.1, LTP:44.17)& 0 & 0 & 0 \\
5 (VTG:0.1, LTP:46.38)& 0 & 0 & 0 \\
6 (VTG:0.1, LTP:49.59)& 0 & 0 & 0 \\
7 (VTG:0.2, LTP:44.17)& 0 & 0 & 0 \\
8 (VTG:0.2, LTP:46.38)& 0 & 0 & 0 \\
9 (VTG:0.2, LTP:49.59)& 0.98 & 0.19 & 0.53 \\ \hline
Expected overall & 0.06 & 0.01 & 0.03 \\ \hline 
\multicolumn{4}{c}{Option 3} \\ \hline
Scenario & Queue & pQueue & Angry \\
1 (VTG:0, LTP:44.17) & 0 & 0 & 0 \\
2 (VTG:0, LTP:46.38) & 0.01 & 0.003 & 0.008 \\
3 (VTG:0, LTP:49.59) & 0.57 & 0.11 & 0.31 \\
4 (VTG:0.1, LTP:44.17)& 0.54 & 0.14 & 0.31 \\
5 (VTG:0.1, LTP:46.38)& 5.03 & 1.24 & 2.85 \\
6 (VTG:0.1, LTP:49.59)& 13.01 & 3.05 & 7.28 \\
7 (VTG:0.2, LTP:44.17)& 11.74 & 3.5 & 6.8 \\
8 (VTG:0.2, LTP:46.38)& 17.81 & 5.14 & 10.26 \\
9 (VTG:0.2, LTP:49.59)& 22.18 & 6.19 & 12.71 \\ \hline
Expected overall & 6.39 & 1.72 & 3.65 \\ \hline
\end{tabular}
\end{table}

\begin{table*} \centering
\caption{The scoring of the different criteria.}
\label{summ_score}
\begin{tabular}{cccccc}
&Criteria & Type & Option 1 & Option 2 & Option 3 \\ \hline \hline
\multirow{4}{*}{Costs (C.1-C.4)} & Environmental Costs & Monetary & (38395.3) & (12798.5) & (12798.5) \\
&Facility Building & Monetary & (0) & (92148.8) & (92148.8) \\
&Additional Safety & Monetary &(205146.8) & (205146.8) &(205146.8) \\\
&Cost Total & Monetary & (243542.1) & (310 094.1) & (310 094.1) \\ \hline \hline
Benefit (B.1) & Additional traffic profit & Monetary & 758 800 & 758 800 & 758 800 \\ \hline \hline
\multirow{2}{*}{Benefit (B.3)} &Local profits & Non-monetary & Yes & Yes & Yes \\
&Job Opportunities & Non-monetary & No & Yes & Yes \\ \hline \hline 
\multirow{4}{*}{Benefit (B.2)}& Road safety & Non-monetary & Yes & Yes & Yes \\
&Queue frequency & Non-monetary & 9.16\% & 0.06\% & 6.39\% \\
&Passive queue frequency & Non-monetary & 2.64\% &0.01\% & 1.72\% \\
&Customer dissatisfaction & Non-monetary & 5.28\% & 0.03\% & 3.65\% \\ 
\end{tabular}
\end{table*}

\subsection{Determining weights and overall scores} \label{det_weights}
The actual scoring of each criteria for the various option are summarised in Table \ref{summ_score}. The MCDA scoring that maps each criteria score to a comparable scale is summarised in Table \ref{summ_score2}. The scoring is done by relative importance. The binary non-monetary scores (yes or no ) are counted as a score of 100 for a yes and a score of 0 for a no. The monetary scores are calculated based on the relative importance of their value and the non-monetary scores are calculated based on the relative importance of their inverse value. The monetary scores are 100 if they are the optimal value and 0 otherwise. The non-monetary percentages scores are calculated based on the relative importance of their inverse value.
\begin{table*} \centering
\caption{The scoring of the different criteria on equal scales.}
\label{summ_score2}
\begin{tabular}{cccccc}
&Criteria & Type & Option 1 & Option 2 & Option 3 \\ \hline \hline
Costs (C.1-C.4) &Cost Total & Monetary & 100 & 0 & 0 \\ \hline \hline
Benefit (B.1)&Additional traffic profit & Monetary & 100 & 100 & 100 \\ \hline \hline
\multirow{2}{*}{Benefit (B.3)} &Local profits & Non-monetary & 100 & 100 & 100 \\
&Job Opportunities & Non-monetary & 0 & 100 & 100 \\ \hline \hline 
\multirow{4}{*}{Benefit (B.2)}&Road safety & Non-monetary & 100 & 100 & 100 \\
&Queue frequency & Non-monetary & 0 & 100 & 30.4 \\
&Passive queue frequency & Non-monetary & 0 & 100 & 35 \\
&Customer satisfaction & Non-monetary & 0& 100 & 31 \\ 
\end{tabular}
\end{table*}

After discussions with the decision makers and stakeholders it was decided that the cost and profit for the port is the most important, so these are given a weighting of 0.25 each, then the customer satisfaction is important so each of these criteria are given a weighting of 0.1 and finally the local community criteria are given a weighting of 0.05 each. The final dynamic MCDA scores of the options are presented in Table \ref{fin_score}.
\begin{table*} \centering
\caption{The overall weighted score of the different options for the dynamic MCDA.}
\label{fin_score}
\begin{tabular}{ccccc c}
Criteria & Type & Option 1 & Option 2 & Option 3 & Weight \\ \hline \hline
Cost Total & Monetary & 100 & 0 & 0 & 0.25 \\ \hline \hline
Additional traffic profit & Monetary & 100 & 100 & 100 & 0.25 \\ \hline \hline
Local profits & Non-monetary & 100 & 100 & 100 & 0.05 \\
Job Opportunities & Non-monetary & 0 & 100 & 100 & 0.05 \\ \hline \hline 
Road safety & Non-monetary & 100 & 100 & 100 & 0.1 \\
Queue frequency & Non-monetary & 0 & 100 & 30.4 & 0.1 \\
Passive queue frequency & Non-monetary & 0 & 100 & 35 & 0.1\\
Customer satisfaction & Non-monetary & 0& 100 & 31 &0.1 \\ 
\multicolumn{2}{r}{Total :} & 65 & 75 & 54.64 & \\\
\end{tabular}
\end{table*}

\subsection{Examination and sensitivity analysis} \label{sensitivity}
The results show the optimal solution is option 2. As the MCDA construction required numerous assumptions, it is important to apply a sensitivity analysis to ensure the optimal option is stable. In general, a sensitivity analysis is applied by just perturbing the weights of the MCDA and investigating how a small perturbation affects the option ranking. However, as this case study was implemented under a high level of uncertainty, the MCDA scores are likely to have a small amount of error. To determine whether the ranking of the options is stable, we applied a sensitivity analysis where each option score could fluctuated by $\pm 10$\% (down to a minimum score of 0). This was accomplished by multiplying each score by an independent identically distributed uniform value between $0.9-1.1$. Letting the score matrix be represented by $S$, the perturbed score matrix $\hat{S}$ was calculated by,
\begin{equation}
\hat{S}_{ij} = S_{ij}\alpha_{ij}, \; \; \alpha_{ij} \sim U(0.9,1,1)
\end{equation}
At first the following criteria were not fluctuated, as they are considered equal for all options, the additional traffic profit, the local profits and the road safety. Therefore $ \forall i\in \{2,3,4 \}$ $\forall j$ $\alpha_{ij}=1 $. The weights were also kept constant. When only the selected criteria were uniformly perturbed by $\pm 10$\% 10000 times, option 2 was ranked highest all of the time by dynamic MCDA. This perturbation range was chosen based on discussions with the stake holders who considered this the maximum range of flexibility.

Some of the criteria are expected to be equal for each option as they are independent of the option, but in reality there is likely to be some deviation. Due to this we also decided to do an additional sensitivity analysis by perturbing all the criteria 10000 times. In this situation, option 2 was ranked highest 56.5\% of the time, option 1 was ranked highest 31.3\% of the time and option 3 was ranked highest 12.2\% of the time. 

It is also possible that the weightings are incorrect, so a final sensitivity analysis was implemented to investigate perturbing all the criteria and the weights, 
\begin{equation}
\hat{w}_{i} = \frac{w_{i}\beta_{i}}{\sum_{k}{w_{k}\beta_{l}}}, \; \; \beta_{i} \sim U(0.9,1,1)
\end{equation}
This time option 2 was ranked highest 56.6\% of the time, option 1 was ranked highest 31.1\% of the time and option 3 was ranked highest 12.3\% of the time. Overall, the sensitivity analysis shows that option 2 is generally stable.

It is clear from these sensitivity results that option 2 will be the optimal solution unless the weight for the cost is greater than the sum of the weights for the local job opportunities, the queue frequency, the passive queue frequency and the customer satisfaction. If the Board give equal weighting to the cost and profit of the port, then if their weighting is $\geq 0.335$ then option 1 will always be optimal. Option 1 would also be more optimal if local profits were considered more important than local job opportunities and road safety was considered more important than queuing and customer satisfaction.

\section{Results \& Discussion} \label{des}
The results show that based on the dynamic MCDA the optimal decision is option 2, to expand the port by building a new weighbridge as this option dominates the option of building a new non-lorry lane at the weighbridge point and has the benefit of improving the customer satisfaction relative to the option of no expansion. However, if the weighting were changed so the the cost was considered the main factor, then option 1 could be more suitable.

Interestingly, if the simulation was not used and the customer probability of queuing, passive queuing and having a poor temperament were not considered, then the scores for option 1-3 would be 93, 64 and 64 respectively. In this case, option 1 would be the optimal decision and this would only change if a large weight was added to the benefit of jobs in the local community, which is very unlikely. Similarly, if only the monetary values were used for a CBA, then option 1 would be optimal. This shows that the incorporation of simulation into the MCDA that enables the inclusion of dynamic criteria can change the final decision. 

For the long term, customer satisfaction is important as this criterion is likely to influence the future profits and unhappy customers are unlikely to use the port again. Therefore, using simulation to aid MCDA may enable decisions to be made based on the long term. This is an important result as many decision problems would want to avoid short sightedness and even though the simulation adds complexity it could be highly beneficial.

\section{Conclusions} \label{conc}
In this paper we have presented a case study to demonstrate a method of using simulation to enable the incorporation of uncertain dynamic criteria into MCDA. We compared the results obtained by applying a dynamic MDCA, a static MCDA and CBA to investigate the optimal expansion solution for the port of Dover when considering service, safety and profit.  The dynamic MDCA returned unique results when compared with the static MCDA or CBA, whereas the static MCDA and CBA returned the same result. The dynamic MCDA that implements a simulation to score novel dynamic criteria may enable decisions to be made that are long term but adds complexity to the process. Nonetheless, this could be beneficial to policy makers when interested in long term goals.

There seems to be a current problem of short sightedness for decision making in the UK and the simulation aided MCDA presents a method of dealing with this.
Therefore possible areas of future work include validating the ideas across more scenarios and different simulations or making the objects more active.

%
\bibliographystyle{kluwer}
\bibliography{dover_ref}
%
%

\end{document}